# Cross-Modal Contrastive Learning from Histopathology and CT for Automated Renal Cell Carcinoma Grading

Amit Das[1], Tanmay Shukla[2], Naofumi Tomita[2], Faraz Farhadi[3], Jessica Sin[4], Ari Hakimi[5], Chad Vanderbilt[6], Jie-Fu Chen[6], Ritesh Kotecha[7], Weijie Ma[8], Bing Ren[8], Saeed Hassanpour[1,2,9*]

1. Department of Computer Science, Dartmouth College, Hanover, NH 03755, USA

2. Department of Biomedical Data Science, Geisel School of Medicine at Dartmouth, Hanover, NH 03755, USA

3. Department of Radiology, Massachusetts General Hospital, Boston, MA 02114, USA

4. Department of Radiology, Dartmouth-Hitchcock Medical Center, Lebanon, NH 03756, USA

5. Department of Urology, Memorial Sloan Kettering Cancer Center, New York, NY 10065, USA

6. Department of Pathology, Memorial Sloan Kettering Cancer Center, New York, NY 10065, USA

7. Genitourinary Oncology Service, Division of Solid Tumor Oncology, Department of Medicine, Memorial Sloan Kettering Cancer Center, New York, NY 10065, USA

8. Department of Pathology and Laboratory Medicine, Dartmouth-Hitchcock Medical Center, Lebanon, NH 03756, USA

9. Department of Epidemiology, Geisel School of Medicine at Dartmouth, Hanover, NH 03755, USA

Corresponding author: Saeed Hassanpour

Email: Saeed.Hassanpour@dartmouth.edu

Postal address: One Medical Center Drive, HB 7261, Lebanon, NH 03756, USA

Phone: (603) 646-5715

## TAKE HOME MESSAGE

A pathology-guided contrastive learning framework improved CT-based classification of low- versus high-grade clear cell renal cell carcinoma while requiring only CT at inference, enabling practical, noninvasive preoperative risk stratification.

**ABSTRACT**

**Background:** Clear cell renal cell carcinoma (ccRCC) exhibits substantial clinical heterogeneity, and accurate grade assessment is essential for risk stratification and treatment planning. However, conventional grading requires invasive tissue sampling. We developed RCC-Align, a cross-modal contrastive learning framework that leverages paired histopathology and computed tomography (CT) data during training to improve noninvasive CT-based ccRCC grade prediction.

**Methods:** RCC-Align aligns paired whole-slide histopathology images (WSIs) and CT scans through contrastive cross-modal objectives, transferring grade-discriminative information from microscopic tissue morphology to macroscopic radiologic representations. The framework was trained and evaluated on paired TCGA and CPTAC cohorts using patient-level five-fold cross-validation. Performance for low- versus high-grade ccRCC classification was compared against CT-only baselines (DINOv2-Base and DINOv2-Finetuned) and a WSI-based reference model (GigaPath-Finetuned). Cross-modal alignment was assessed using cosine similarity analysis.

**Results:** RCC-Align achieved an AUC of 0.601 (95% CI, 0.524-0.673) and AUPRC of 0.599 (95% CI, 0.541-0.676), outperforming DINOv2-Finetuned (AUC 0.545; AUPRC 0.543) with significantly improved low-grade prediction ($p = 0.004$). RCC-Align also demonstrated stronger paired WSI-CT embedding alignment compared with baselines. The WSI-based GigaPath reference achieved an AUC of 0.719.

**Conclusion:** Pathology-guided contrastive learning improves CT-based ccRCC grading while requiring only CT at inference. This approach may complement tissue diagnosis when biopsy is unsafe, infeasible, or limited by intratumoral heterogeneity. Validation in larger, multi-institutional cohorts with external testing is needed before clinical translation.

## INTRODUCTION

Renal cell carcinoma (RCC) is the most common malignancy of the kidney, and clear cell renal cell carcinoma (ccRCC) accounts for 70-80% of RCC cases.[1] Because ccRCC exhibits marked clinical heterogeneity, ranging from indolent localized tumors to aggressive metastatic disease, accurate risk stratification is essential for treatment planning and prognostication.[1] Histopathological grade is a key determinant of outcome, and the International Society of Urological Pathology grading system has strong prognostic relevance in ccRCC.[2,3] Reliable preoperative or early postoperative assessment of tumor grade could therefore inform surgical decision-making, surveillance, and selection of systemic or adjuvant treatment strategies.

Current grading relies on tissue sampling, but biopsy-based assessment has important limitations. A single core sample may not capture the substantial intratumoral heterogeneity of ccRCC and can therefore underestimate or misclassify tumor grade.[4] Biopsy also adds procedural risk, cost, and patient burden. In contrast, computed tomography (CT) is routinely obtained for the detection and staging of renal masses, is widely available, and is noninvasive, but conventional CT alone does not reliably provide the tumor grade information needed for clinical decision-making. A more accurate imaging-based approach could therefore complement tissue diagnosis, particularly in settings in which biopsy is unsafe, infeasible, or limited by sampling variability.

Recent deep learning studies have shown that both whole-slide histopathology and radiologic imaging contain useful information for RCC characterization and grade prediction.[5,6] Multimodal approaches have further demonstrated promise in ccRCC, but most prior work has focused on prognosis, recurrence, or survival rather than histopathological grade classification.[7–10] Although one prior study addressed grade prediction, it relied on radiomics and transcriptomic data rather than an imaging-only deep learning framework.[11] More broadly, many existing multimodal approaches depend on hand-crafted features, genomic inputs, or simple fusion strategies, which may limit scalability and clinical applicability.[7–11] These gaps highlight the need for methods that can directly learn cross-scale associations between microscopic tissue morphology and macroscopic imaging phenotypes while remaining practical for deployment.

Here, we present RCC-Align, a cross-modal contrastive learning framework that uses paired histopathology whole-slide images and CT scans during training to improve CT-based prediction of low-grade versus high-grade ccRCC. Building on recent advances in pathology foundation models and cross-modality representation learning,[12,13] RCC-Align aligns histopathology and CT representations so that grade-relevant information learned from tissue can improve radiologic prediction. Importantly, although the model is trained using paired multimodal data, it performs inference using CT alone. This design preserves a clinically practical deployment setting while leveraging histopathology supervision during training to enhance noninvasive grade estimation.

## METHODS

### Overview

RCC-Align is a deep learning framework for predicting clear cell renal cell carcinoma grade from CT images by leveraging complementary information from paired histopathology whole-slide images during training. Histopathology features are extracted using the Prov-GigaPath foundation model to generate patch-level embeddings that are aggregated into slide-level representations of global tissue morphology, while CT features are derived using a pre-trained DINOv2 encoder. During training, RCC-Align jointly optimizes a supervised classification loss based on pathologist-assigned grades and a cross-modal contrastive loss that aligns histopathology and CT embeddings.

This design enables CT-based grade prediction at inference while benefiting from multimodal supervision during training. The framework was evaluated on a paired TCGA and CPTAC dataset.

### Outcome definition

Clear cell renal cell carcinoma was dichotomized into low-grade (ISUP grades 1-2) and high-grade (ISUP grades 3-4) disease. This binary classification aligns with clinical decision-making, which often depends on distinguishing indolent from aggressive tumors rather than resolving the full four-tier grading system. Higher ISUP grades are associated with worse oncologic outcomes, supporting the clinical relevance of this grouping.[14–16] Binary stratification also reduces label noise, mitigates class imbalance in limited paired datasets, and improves annotation reliability because interobserver agreement is higher for low- versus high-grade categories than for adjacent individual grades.[14,17] This formulation therefore preserves clinically actionable information while improving statistical robustness and model tractability.

### Datasets for Fine-tuning GigaPath Encoder

RCC-Align was trained using datasets from the Clinical Proteomic Tumor Analysis Consortium (CPTAC), The Cancer Genome Atlas (TCGA), and Memorial Sloan Kettering Cancer Center (MSK). CPTAC and TCGA provided paired H&E whole-slide images and CT scans for ccRCC cases, whereas MSK contributed a large WSI-only ccRCC cohort. The H&E slides were derived from surgical resection specimens rather than needle core biopsies, providing broader sampling of tumor morphology. However, because ccRCC is intrinsically heterogeneous, individual WSIs may not capture the full grade distribution of the tumor and instead reflect the sampled sections selected for pathologic review, which may be enriched for tumor-rich or higher-grade regions. Unpaired WSIs from CPTAC and TCGA were retained for single-modality fine-tuning to expand the histopathology training set despite incomplete cross-modal pairing. Additional dataset details are provided in Table 1.

| Dataset | Modality | # Patients | # WSIs | # CT | Low/High Grade |
|---|---|---|---|---|---|
| TCGA | H&E + CT | 162 | 163 | 162 | 62/101 |
| CPTAC | | 49 | 149 | 49 | 90/59 |
| MSK | H&E Only | 1,179 | 9,641 | — | 3,748/5,893 |
| TCGA | | 326 | 329 | — | 157/172 |
| CPTAC | | 138 | 496 | — | 256/240 |
| DHMC | | 101 | 123 | — | 100/23 |

**Table 1:** Summary of datasets used in this study. For each dataset source, the table lists the modality, total number of available WSIs, and class distribution. High-grade corresponds to ISUP grades 3–4 and low-grade corresponds to ISUP grades 1–2.

*Paired WSI and CT Dataset*

The paired CPTAC and TCGA dataset comprised 312 WSIs from 211 patients. The TCGA subset included 101 high-grade and 62 low-grade cases, and the CPTAC subset included 59 high-grade and 90 low-grade cases. CT labels were assigned at the patient level, with some patients contributing multiple WSIs. Model development used five-fold cross-validation stratified by patient to prevent data leakage and balanced by cohort and grade.

All CT scans were contrast-enhanced, but contrast phase varied across patients. Based on DICOM metadata, only a minority of scans were explicitly labeled as venous or arterial phase, whereas most were delayed phase or lacked a specific phase label. To reduce variability from this heterogeneity, all scans underwent the same preprocessing workflow, including kidney segmentation, yielding standardized contrast-enhanced CT inputs rather than a uniform single-phase protocol.

*WSI-Only Fine-Tuning Cohort*

The WSI-only dataset included 10,589 WSIs from MSK, TCGA, CPTAC, and Dartmouth-Hitchcock Medical Center, of which 6,328 were high-grade and 4,261 were low-grade. Fine-tuning was initialized from a metastasis-pretrained checkpoint and continued on the MSK WSIs using RCC grade labels. This transfer learning strategy allowed the histopathology encoder to adapt to grade-specific RCC morphology and improve the WSI representations used in RCC-Align.

## Data Preprocessing

The preprocessing pipeline extracted patches from H&E whole-slide images and computed tile-level embeddings for each patch. For CT, we used TotalSegmentator[18] to segment the kidneys and generate kidney masks for downstream processing. The preprocessing details are provided below.

*Patch Extraction & Embedding Generation*

Binary tissue masks were generated from downsampled WSIs and upscaled to 20× magnification.[19] Non-overlapping 224 × 224-pixel patches were then extracted using a sliding window approach from regions containing at least 20% stained tissue. Tile embeddings were generated for each patch using the Prov-GigaPath tile encoder, which was pretrained on 20× H&E and IHC patches of the same size. The resulting embeddings and patch coordinates were stored for downstream analysis.

*Generating Segmentations for CT Images*

Kidney regions were segmented from each CT volume using TotalSegmentator to enable consistent slice-based processing.[18] CT volumes and kidney masks were then loaded with NiBabel,[20] and the left and right kidney masks were merged into a single binary mask. Axial slices containing kidney tissue were retained, windowed, resized to 512 × 512 pixels, and saved for downstream analysis. This preprocessing pipeline enabled consistent extraction of kidney-centered CT slices across patients and reduced background variability for patient-level feature aggregation. The CT preprocessing workflow is shown in Figure 1.

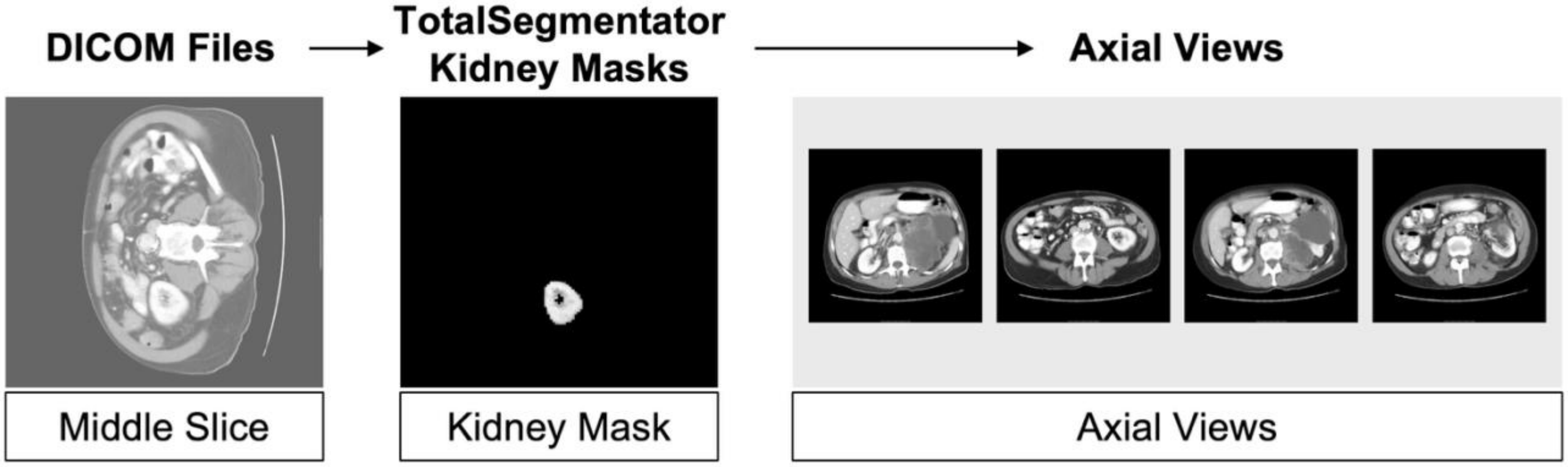

**Figure 1:** CT preprocessing. DICOM CT volumes are segmented with TotalSegmentator to obtain voxel-wise kidney masks. Axial slices are extracted, and only slices containing kidney tissue are retained. The selected slices are windowed, resized, and saved at a fixed resolution of 512×512.

**RCC-Align Framework**

RCC-Align predicts RCC grade from CT images by leveraging paired H&E and CT data during contrastive training. WSI representations are generated by aggregating patch embeddings from the Prov-GigaPath tile encoder with the Prov-GigaPath slide encoder. CT scans are processed as sequences of axial slices, each encoded with a DINOv2-based vision transformer,[21] and slice-level features are max-pooled to obtain a patient-level CT representation. A linear classification head is applied to this pooled CT embedding for grade prediction.

During training, RCC-Align aligns paired WSI and CT representations using a symmetric contrastive loss while preserving grade-discriminative information through modality-specific classification losses. At inference, the model operates in a single-modality setting, using the trained CT encoder and classification head to generate patient-level predictions from CT alone. This design enables CT-based RCC grade prediction while benefiting from multimodal supervision during training. The RCC-Align architecture is shown in Figure 2.

**Figure 2:** RCC-Align architecture. During training, the model optimizes inter-modality, intra-modality, and classification losses. H&E patches are encoded by a ViT-based Patch Encoder to produce tile embeddings, which are aggregated by a LongNet-based Slide Encoder into slide-level representations. Patient CT scans are treated as sequences of axial slices, each encoded with a DINOv2 ViT backbone; features are averaged per slice and then max-pooled across slices to form a patient-level CT embedding. A contrastive loss aligns H&E and CT representations across

modalities. Masked H&E tiles are used to compute the intra-modality loss, while the classification loss supports the downstream prediction task.

*Loss Functions*

RCC-Align is trained with three complementary objectives: an inter-modality contrastive loss, an intra-modality consistency loss, and modality-specific classification losses. Together, these losses align histopathology and CT representations while preserving information relevant to ccRCC grade prediction. The inter-modality loss aligns paired histopathology and CT representations from the same patient. The inter-modality loss encourages alignment between paired histopathology and CT representations derived from the same patient. Let $q_i \in \mathbb{R}^d$ denote the WSI embedding for the $i$-th sample and ${k_i}^+ \in \mathbb{R}^d$ denote the corresponding CT embedding. RCC-Align uses a symmetric contrastive loss based on the InfoNCE formulation, adapted from the TANGLE framework[22], to bring paired WSI-CT embeddings closer while separating unmatched samples within a batch. For a batch of N paired samples, the inter-modality loss is defined as:

$$L_{inter} = \frac{1}{2N}\sum_{i=1}^{N} \left[CE\left(\frac{q_i \cdot K^+}{\tau}, y_i\right) + CE\left(\frac{{k_i}^+ \cdot Q}{\tau}, y_i\right)\right]$$

where $Q \in \mathbb{R}^{N\times d}$ denotes the matrix of WSI embeddings formed by stacking all $q_i$ vectors in the batch, $K^+ \in \mathbb{R}^{N\times d}$ denotes the corresponding matrix of CT embeddings formed by stacking all $k_i^+$ vectors, $\tau$ is the softmax temperature, and CE denotes the cross-entropy loss computed over similarity scores. Positive pairs correspond to embeddings from the same patient.

To improve robustness within the histopathology modality, RCC-Align also uses an intra-modality consistency loss based on masked embedding reconstruction. During training, 50% of patch-level embeddings within each WSI are randomly masked. The masked and unmasked patch sequences are passed through the same slide encoder, producing embeddings $Z$ and $Z_m$, respectively. The intra-modality loss minimizes the mean squared error between these embeddings:

$$L_{intra} = L_{MSE}(Z, Z_m)$$

To preserve discriminative information for grading, supervised classification losses are applied to both modalities. Patient-level CT embeddings and slide-level WSI embeddings are passed through their respective classification heads to predict RCC grade. Given logits ${\hat{y}_i}^{CT}$ and ${\hat{y}_i}^{WSI}$ and ground truth label $y_i$, the classification loss is:

$$L_{class} = CE({\hat{y}_i}^{CT}, y_i) + CE({\hat{y}_i}^{WSI}, y_i)$$

The overall training objective is the unweighted sum of the three components:

$$L_{total} = L_{inter} + L_{intra} + L_{class}$$

By jointly optimizing these objectives, RCC-Align learns modality-aligned representations that remain predictive of RCC grade. At inference, the framework supports single-modality deployment, enabling CT-only or WSI-only prediction while benefiting from multimodal supervision during training.

**Evaluation**

RCC-Align was evaluated using five-fold nested cross-validation with 60%/20%/20% training, validation, and test splits. All slides from the same patient were assigned to the same fold to prevent data leakage. Performance was assessed using area under the receiver operating characteristic curve (AUC), F1 score, precision, and recall, with results aggregated across test folds. Confusion matrices were generated, and 95% confidence intervals were estimated by bootstrapping.

To assess the contribution of cross-modal and intra-modality objectives, a baseline model trained using only the classification loss, termed DINOv2-Finetuned, was evaluated using the same procedure. Cosine similarity between paired and unpaired H&E and CT embeddings was also computed for DINOv2-Finetuned and RCC-Align to assess cross-modal alignment. DINOv2-Base was included as an additional CT baseline. Finally, GigaPath was fine-tuned on 10,589 unpaired WSIs from MSK, TCGA, CPTAC, and DHMC and evaluated on WSIs from the paired cohort to provide a histopathology-based reference for grading performance when WSI data are available.

## RESULTS

### Model Performance

RCC-Align, which was trained using paired H&E whole-slide images and CT scans but evaluated using CT alone, was compared with DINOv2-Finetuned, which was trained using only the CT classification loss and did not use the paired structure of the dataset. Results for DINOv2-Base, DINOv2-Finetuned, and RCC-Align are shown in Table 2. RCC-Align significantly outperformed DINOv2-Finetuned for low-grade tumor prediction on Wilcoxon signed-rank testing of prediction scores ($p = 0.004$). As expected, RCC-Align performed worse than GigaPath-Finetuned, which uses whole-slide images that provide richer and higher-resolution information for grading.

| Task | Input modality | Model | AUC | F1 | AUPRC |
|---|---|---|---|---|---|
| RCC Grade Prediction | *Whole-slide Images (WSIs)* | *GigaPath-Finetuned* | *0.719 [0.661-0.774]* | *0.653 [0.599-0.704]* | *0.696 [0.644-0.759]* |
| | Computed Tomography Images (CTs) | DINOv2-Base | 0.531 [0.492-0.644] | 0.519 [0.454-0.583] | 0.530 [0.479-0.603] |
| | | DINOv2-Finetuned | 0.545 [0.453-0.601] | 0.542 [0.449-0.582] | 0.543 [0.479-0.600] |
| | | **RCC-Align** | **0.601 [0.524-0.673]** | **0.564 [0.498-0.630]** | **0.599 [0.541-0.676]** |

**Table 2:** Performance comparison of multiple frameworks across the CT RCC Grade classification task. Evaluated models include baseline methods such as DINOv2-Base (pre-trained DINOv2 encoder), DINOv2-Finetuned (fine-tuned DINOv2 encoder), and RCC-Align (fine-tuned CT encoder using inter-modality, intra-modality, and class alignment losses). GigaPath-Finetuned denotes the WSI-trained GigaPath slide encoder, which serves as the histopathology reference classification and embeddings and provides the inter-modality alignment target for training RCC-Align. Metrics reported are AUC, weighted F1-score, and Area Under Precision Recall Curve (AUPRC), with 95% confidence intervals provided in brackets. **Bold** values indicate the best performance for each metric in the CT-based evaluation.

**Cosine Similarity Analysis**

DINOv2-Finetuned trains the DINOv2 encoder using only the classification loss, whereas RCC-Align incorporates inter- and intra-modality losses in addition to the classification loss. To assess cross-modal alignment, cosine similarity between WSI and CT embeddings was computed for paired H&E-CT samples in the test set and averaged across folds. For each H&E WSI, cosine similarity was also computed against the remaining CT scans in the test set to generate random mismatched pairs. As shown in Table 3, RCC-Align produced higher cosine similarity for paired than for unpaired WSI-CT pairs, and this difference was more pronounced than for DINOv2-Finetuned. These findings indicate that RCC-Align improves alignment between histopathology and CT representations.

| Model | Paired Cosine Similarity (A) | Shuffled Cosine Similarity (B) | Difference (A-B) |
|---|---|---|---|
| DINOv2-Base | -0.024 | -0.025 [-0.031– -0.019] | 0.000 [0.000–0.000] |
| DINOv2-Finetuned | -0.001 | -0.001 [-0.007–0.004] | 0.000 [0.000–0.000] |
| RCC-Align | 0.092 | 0.088 [0.082–0.094] | 0.003 [0.003–0.004] |

**Table 3:** Cross-modality alignment analysis comparing DINOv2-Finetuned and RCC-Align models on the classification task. Paired Cosine Similarity: Computed between embeddings of correctly matched H&E WSIs and CT images and averaged across cross-validation folds. Shuffled Cosine

Similarity: Computed by repeatedly shuffling WSI–CT pairings within each fold and reported as mean ± 95% confidence interval across shuffles. Difference: Computed as the fold-wise difference between paired and shuffled similarities; mean ± 95% CI is reported across shuffled repetitions. A Wilcoxon signed-rank test showed that RCC-Align achieves significantly stronger alignment than DINOv2-Finetuned by comparing per-slide difference scores.

**DISCUSSION**

RCC-Align demonstrated that cross-modal contrastive learning using paired histopathology and CT data during training can improve noninvasive CT-based prediction of ccRCC grade compared with CT-only baselines. The framework achieved this improvement by aligning radiologic representations with grade-discriminative features extracted from a pathology foundation model, while requiring only CT at inference. These results support the hypothesis that microscopic tissue morphology, when used as a supervisory signal during training, can enrich the information captured in macroscopic imaging embeddings.

The observed improvement in cross-modal alignment between paired WSI and CT embeddings, together with the significant gain in low-grade prediction accuracy, suggests that explicit cross-modal supervision provides a learning signal beyond what classification loss alone can achieve. This finding has implications beyond ccRCC grading. Paired histopathology and radiology datasets, though less common than single-modality collections, are increasingly available through institutional archives and public consortia such as TCGA and CPTAC. The contrastive alignment strategy demonstrated here could be extended to other tumor types and imaging settings in which tissue-level features carry prognostic information that is not fully captured by standard radiologic assessment.

Clinically, CT-based grade prediction may serve as a front-end risk stratification tool for patients with known or suspected ccRCC, particularly when biopsy is unsafe, infeasible, or limited by sampling variability and intratumoral heterogeneity. Rather than replacing tissue diagnosis, this approach is best understood as an adjunct that leverages imaging already acquired in routine clinical care. Renal mass biopsy can be informative but has recognized limitations for grading, as a single sampled region may not capture the full biologic heterogeneity of the tumor. At the same time, a substantial proportion of small renal masses are indolent, and active surveillance is an accepted management strategy in selected patients. In this context, an imaging-based estimate of histologic grade could provide a complementary signal that supports surveillance more confidently in likely low-grade disease while directing higher-risk cases earlier toward biopsy or definitive treatment.

Several limitations should be acknowledged. The paired cohort was relatively small, imaging protocols were heterogeneous across institutions, and the CT encoder was not pretrained on domain-specific CT data. RCC-Align also performed below the histopathology-based GigaPath reference model, which reflects the inherent information asymmetry between macroscopic cross-sectional imaging and microscopic whole-slide histopathology. However, because RCC-Align benefits from paired CT-WSI supervision only during training while remaining CT-only at inference, its performance ceiling is not fixed; larger and more diverse multi-institutional training sets, CT-specific encoder pretraining, and incorporation of contrast-enhanced imaging phases could each contribute to further gains. Although CT was a logical initial modality given its broad availability and routine role in renal mass evaluation, MRI represents an important future direction. Its superior soft-tissue contrast and growing clinical use in renal mass characterization may provide additional predictive signal for imaging-based grade estimation.

In summary, RCC-Align establishes that pathology-guided contrastive learning can transfer grade-discriminative information from microscopic tissue morphology to macroscopic radiologic representations, enabling improved noninvasive classification of low- versus high-grade ccRCC from CT alone. These findings position cross-modal alignment as a practical strategy for enriching imaging-based tumor characterization without altering the clinical workflow at deployment. Validation in larger, multi-institutional cohorts with external test sets and prospective clinical evaluation will be essential to define the clinical utility of this approach.

## Declarations

### Acknowledgements

The authors thank Tracy Frazee and Manu Goyal for their valuable help during the study.

### Ethics approval and consent to participate

This study was conducted in accordance with the Declaration of Helsinki (as revised in 2013) and was approved by the Institutional Review Board of Dartmouth Health (IRB ID: MOD00027730; approved December 9, 2025) under expedited review. The study was classified as no greater than minimal risk, and waivers of consent documentation and HIPAA authorization were granted given the retrospective design and use of existing data, documents, records, and specimens. Publicly available de-identified data from The Cancer Genome Atlas (TCGA) and the Clinical Proteomic Tumor Analysis Consortium (CPTAC) were used in accordance with their respective data access policies.

### Consent for publication

Not applicable.

**Competing interests**

The authors declare that they have no competing interests.

**Availability of data and materials**

The CPTAC and TCGA datasets are publicly available and accessible from their corresponding publishers. The DHMC and MSK datasets generated and analyzed during this study may be made available from the corresponding author upon reasonable request and with appropriate institutional approvals, subject to patient privacy regulations and institutional data sharing policies.

**Funding**

This research was supported in part by the National Library of Medicine (R01LM013833) and the National Cancer Institute Cancer Center Support Grant to Memorial Sloan Kettering Cancer Center (P30CA008748). R.R.K. is supported in part by a Department of Defense Kidney Cancer Research Program Early Career Investigator Award (W81XWH-21-1-0942) and a Focus Award from the Kidney Cancer Association in partnership with Joey's Wings Foundation.

**Authors' contributions**

Conceptualization: F.F. and S.H.; Methodology: A.D., T.S., and S.H.; Software: A.D.; Formal analysis: A.D. and T.S.; Investigation: A.D., T.S., N.T., B.R., and S.H.; Resources: S.H.; Data curation: F.F., J.M.S., A.H., C.V., J.-F.C., R.R.K., W.M., and B.R.; Writing, original draft: A.D.; Writing, review and editing: all authors; Visualization: A.D. and N.T.; Supervision: S.H.; Funding acquisition: S.H. All authors reviewed and approved the final manuscript.

**Declaration of Generative AI and AI-assisted Technologies in the Writing Process**

During the preparation of this work, the authors used ChatGPT (OpenAI, San Francisco, CA) for language editing and proofreading to improve clarity and readability. The tool was not used for study design, data collection, data analysis, data interpretation, or generation of scientific content.

After using this tool, the authors reviewed and edited all output as needed and take full responsibility for the content of this publication.